%% file: root.tex
\documentclass[letterpaper, 10 pt, conference]{ieeeconf}  

\IEEEoverridecommandlockouts                              

\usepackage{graphicx} 
\usepackage{amsmath} 
\usepackage{amssymb}  
\usepackage{xcolor}
\usepackage{cite}

\usepackage{etoolbox}
\makeatletter
\patchcmd{\@makecaption}
  {\scshape}
  {}
  {}
  {}
\makeatother

\title{\LARGE \bf
Domain-Transferred Synthetic Data Generation for Improving Monocular Depth Estimation
}

\author{Seungyeop Lee$^{1}$, Knut Peterson$^{2}$, Solmaz Arezoomandan$^{2}$, Bill Cai$^{2}$, Peihan Li$^{2}$, \\ Lifeng Zhou$^{2}$, and David Han$^{2}$
\thanks{$^{1}$Affiliated with Korea University  $^{2}$Affiliated with Drexel University}
}

\begin{document}

\maketitle
\thispagestyle{empty}
\pagestyle{empty}

\begin{abstract}
A major obstacle to the development of effective monocular depth estimation algorithms is the difficulty in obtaining high-quality depth data that corresponds to collected RGB images. Collecting this data is time-consuming and costly, and even data collected by modern sensors has limited range or resolution, and is subject to inconsistencies and noise. To combat this, we propose a method of data generation in simulation using 3D synthetic environments and CycleGAN domain transfer. We compare this method of data generation to the popular NYU-Depth V2 dataset by training a depth estimation model based on the DenseDepth structure using different training sets of real and simulated data. We evaluate the performance of the models on newly collected images and LiDAR depth data from a Husky robot to verify the generalizability of the approach and show that GAN-transformed data can serve as an effective alternative to real-world data, particularly in depth estimation. 
\end{abstract}

\input{1.Introduction}

\input{2.Related}

\input{3.Proposed}
\input{4.Experiment}

\input{5.Conclusion}






\section*{ACKNOWLEDGMENT}
The funding for this work was provided by the US Federal Aviation Administration (FAA) under the task A51 of the Alliance for System Safety of UAS through Research Excellence (ASSURE) program.
This research was also supported by the MOTIE (Ministry of Trade, Industry, and Energy) in Korea, under the Fostering Global Talents for Innovative Growth Program related to Robotics(P0017311) supervised by the Korea Institute for Advancement of Technology (KIAT).


\bibliographystyle{IEEEtran}
\bibliography{ref.bib}
\end{document}

%% file: 1.Introduction.tex
\section{Introduction}

Obstacle detection and avoidance are essential for autonomous systems, particularly in applications such as autonomous driving and robotics~\cite{prabhakar2017obstacle,khosravian2021generalizing,quirynen2020integrated,kanchana2021computer}. These systems require accurate distance measurements between robots and nearby obstacles to ensure safe and effective control. While Stereo cameras and LiDAR distance sensors are capable of obtaining depth information, their weight, energy consumption and resolution limitations make them less applicable to small robots constrained by size and power capacities. Monocular cameras have gained prominence in the realm of Micro Aerial Vehicles (MAVs) and mobile robots due to their low cost, lightweight, and ability to capture high-resolution images~\cite{forster2014appearance, forster2015continuous, an2021real, miclea2021monocular, yang2018monocular, weiss2013monocular, wofk2019fastdepth}. However, given that monocular vision inherently lacks depth perception, developing effective methods to estimate distance solely from monocular camera images holds significant potential for a wide range of applications. 

The most effective methods for monocular depth estimation have been driven by developments in deep learning algorithms~\cite{DenseDepth,r7,r10}. Deep learning requires large training datasets to be effective, and supervised learning depth estimation needs both scene images and corresponding depth data collected by precise depth sensors. There are several obstacles to obtaining accurate depth information, such as the low resolution of LiDAR data or the limited range and sensitivity to noise of stereo cameras. Additionally, manual data collection is time-consuming and resource-intensive. A possible alternative can be found in game engines such as Unreal Engine or Unity, which are employed to create synthetic environments that mimic real-world conditions. These synthetic environments allow the generation of images and the corresponding depth values automatically.

While images generated by Unreal Engine or Unity may appear realistic to the naked eye, it has been well acknowledged that they may contain subtle artifacts leading to biases in distance estimations. To bridge this domain gap between synthetic and real world imagery, we propose leveraging Generative Adversarial Network(GAN) for domain transfer ~\cite{goodfellow2014generative}. By utilizing a type of GAN called ``CycleGAN," we transform synthetic images generated in a simulated environment into images closely resembling real-world scenes ~\cite{CycleGAN2017}. Our research uses the transformed images as input for a depth estimation model, demonstrating that training with a transferred synthetic dataset effectively minimizes the disparity between synthetic and real-world data, thereby enhancing depth map accuracy. Our contributions in this work are:
\begin{itemize}
  \item Development of an augmentation method for generating image-depth map pairs for effectively training a depth estimation algorithm.
  \item  An adaptation of CycleGAN for domain transfer between simulated and realistic looking images effective in improving monocular depth estimation.
  
\end{itemize}

%% file: 2.Related.tex
\section{Related Work}

\subsection{Domain Transfer}
 Generative Adversarial Networks (GANs) ~\cite{goodfellow2014generative} have emerged as a powerful class of artificial intelligence algorithms capable of generating realistic-looking images. A GAN consists of two main components: a generator and a discriminator. These components engage in an adversarial process, where the generator aims to produce images that are indistinguishable from real images, effectively ``fooling" the discriminator, while the discriminator's task is to distinguish between genuine and generated images. This concurrent training process enhances the capabilities of both components.

Building on the GAN framework, Cycle-Consistent GAN (CycleGAN) ~\cite{CycleGAN2017} introduces an innovative approach to unsupervised image-to-image translation between two distinct domains using unpaired image datasets. The model operates with two pairs of generators and discriminators, one for each domain. It leverages adversarial loss to train the network, enabling the generators and discriminators to improve through mutual feedback. A distinctive feature of CycleGAN is its use of cycle consistency loss, which ensures that an image can be translated from one domain to another and then back to the original domain without losing its original identity. This feature makes CycleGAN particularly valuable for applications where paired domain data is scarce or unavailable.

The CycleGAN-based method for data augmentation has been recently applied to various tasks in computer vision. In ~\cite{r4}, the domain transferred training datasets improved the detecting performance of YOLO for multi-organ detection in medical images. Also, CycleGAN has been applied to real-to-virtual domain transfer for visualizing transnasal surgery~\cite{r5}. We propose that using CycleGAN to translate simulated images into more realistic images can increase the viability of using simulated data for monocular depth estimation.

\subsection{Monocular Depth Estimation}
Monocular Depth Estimation (MDE), a method for predicting depth information from a single RGB image, has been widely researched, particularly for its applications in lightweight robotics and MAVs. The characteristic that MDE only uses RGB cameras without any other sensor offers a cost-effective and lightweight solution for depth perception.

Early MDE relied on depth cues for depth prediction with strict requirements~\cite{r12}. However, the advent of deep learning led to rapid developments of depth estimation algorithms that outperform traditional methods. 

Supervised learning-based MDE models are trained using datasets that have RGB and corresponding grayscale depth map image pairs such as NYU-Depth V2~\cite{r8} and KITTI~\cite{r9}. Eigen~\cite{r7} first proposed a deep learning-based MDE model which uses scale-invariant error. It has a coarse-to-fine framework, where the coarse network trains the global depth information and the fine network obtains local features. Since then, a variety of deep learning methods for monocular depth estimation emerged~~\cite{DenseDepth,r10,r11}.

The most common architecture for monocular depth estimation based on deep learning is an encoder-decoder network, using RGB images as an input and extracted depth maps as an output. The encoder network consists of convolutional and pooling layers to learn features of images and the decoder network consists of deconvolutional layers to predict depth maps~\cite{r12}. Because of the difficulty of collecting ground truth depth data, several different approaches have been explored to improve either the efficiency of data usage or expand the type of training data that can be used. Ranftl et al.~\cite{r6} suggested several methods to allow training on multiple types of depth datasets, such as proposing training objectives that are invariant to depth range and scale, and exploring multi-objective learning when combining datasets.

While supervised methods have been widely explored, more recently unsupervised and semi-supervised methods have also begun to gain traction, mainly because of their ability to learn without the same level of dependence on ground truth depth data. These methods can learn depth estimation using image datasets without labels, but require significantly more data compared to supervised methods~\cite{r13}. Learning approaches like self-attention first proposed by ~\cite{caron2021emerging} have since been expanded to apply to depth estimation tasks, yielding competitive results~\cite{oquab2023dinov2}. Most notably, recent benchmarks have been set on the NYU Depth V2 dataset by models such as VPD~\cite{Zhao_2023_ICCV} which leverages cross-attention maps between images and text labels to aid segmentation and depth estimation tasks, and Depth Anything~\cite{depthanything} which uses combinations of ground truth labels and pseudo-labels generated in a teacher-student architecture to learn from both labeled and unlabeled image datasets. While both of these models mainly rely on large training image datasets that do not contain depth information, they still require ground truth depth data for fine-tuning.

%% file: 3.Proposed.tex
\section{Proposed Method}
Our proposed method consists of three modules: synthetic data generation, data domain transformation, and depth estimation model training. CycleGAN is used for the domain transformation method to translate synthetic images to realistic-looking images and DenseDepth is used for the depth estimation model.

\subsection{Synthetic Data Generation}

\begin{figure}[t]
    \includegraphics[width=0.5\textwidth]{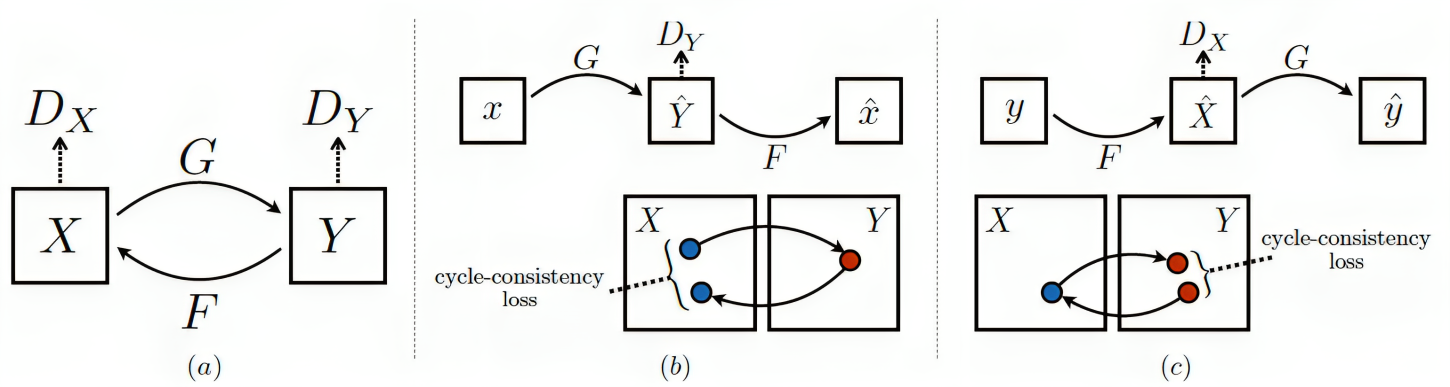}
    \caption{Illustration of (a) CycleGAN mapping functions, (b) forward cycle-consistency, and (c) backward cycle-consistency ~\cite{CycleGAN2017}.}
    \label{fig:GANdiagram}
\end{figure}

\begin{figure*}
    \centering
    \centerline{\includegraphics[width=\textwidth]{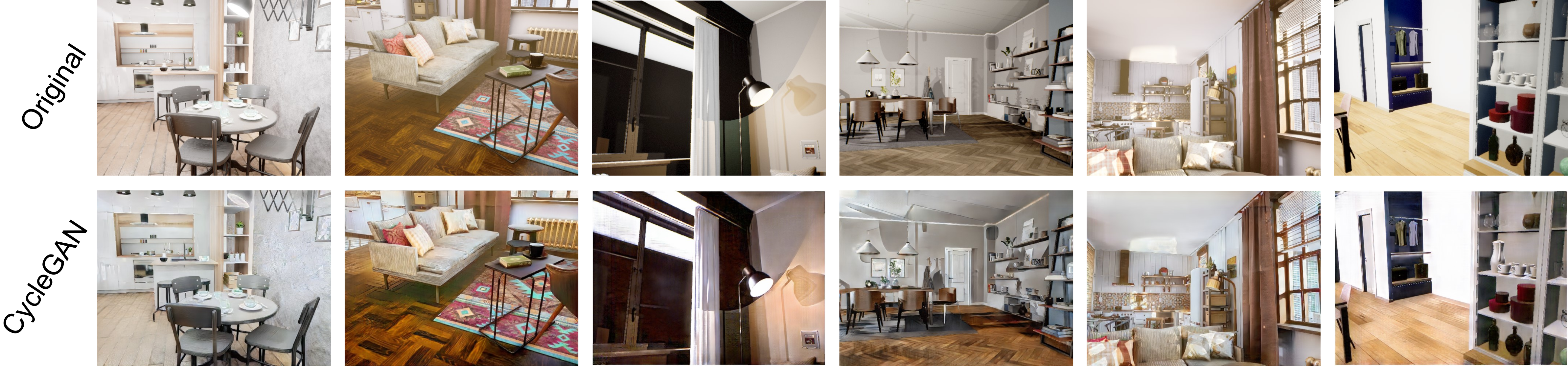}}
    \caption{Synthetic images generated using Unreal Engine (first row) and their CycleGAN-transformed counterparts (second row). The CycleGAN transformation process alters features such as lighting and texture to increase the realism of a simulated image while maintaining the original features of the environment.}
    \label{fig:GANresults}
\end{figure*}

To generate the synthetic image dataset, we utilized the Unreal Engine v4.27 simulation software developed by Epic Games \cite{unrealengine427}, in conjunction with the AirSim \cite{shah2017airsim} for capturing RGB images and pixel-wise distance data. We selected a diverse set of indoor virtual environments, including several apartment scenes, living room, bedroom, and kitchen environments. To match the image data in the NYU-Depth V2 dataset, the RGB images were captured at a resolution of 640x480 pixels, and the field of view angle of the camera was set to 57 degrees. The depth information was captured using the AirSim DepthPerspective camera setting and was saved to the \texttt{.pfm} file format to preserve raw distance information. This \texttt{.pfm} file was then converted to an 8-bit depth image by first truncating any distances exceeding 10 meters to a value of 10 meters, and then linearly scaling the distance values from 0 to 255 and saving the result to a grayscale \texttt{.png} image.

\subsection{Domain Transfer}
CycleGAN's capability for unpaired image-to-image translation is crucial for domain transfer tasks where paired examples are scarce or unavailable. This characteristic aligns with our project's needs, as we aim to translate images from the synthetic domain to the real domain without explicit one-to-one mappings.

As we observe from Figure \ref{fig:GANdiagram}, the architecture of CycleGAN ~\cite{CycleGAN2017} employs two generative adversarial networks (GANs), one for each domain, which are trained simultaneously. The generators aim to transform images from one source domain to another target domain, while the discriminators strive to distinguish between real and generated images. It is important to mention that the terms, terminologies, and equations presented below are directly sourced from ~\cite{CycleGAN2017}.

The objective is to acquire a mapping \(G: X \rightarrow Y\) in a manner that the image distribution generated by \(G(X)\) is indistinguishable from the distribution \(Y\), achieved through an adversarial loss. Given the inherent lack of constraints in this mapping, an inverse mapping \(F: Y \to X\) is integrated to incorporate a cycle consistency loss. This ensures that the process \(F(G(X))\) results in images equivalent to \(X\) (and vice versa), thereby adding a constraint to the otherwise unconstrained mapping. 

In conjunction with the generators, two adversarial discriminators, \(D_{\text{\scriptsize X}}\) and  \(D_{\text{\scriptsize Y}}\), are used in the respective GANs. \(D_{\text{\scriptsize X}}\) is tasked with distinguishing between the original images \(\{x\}\) and their translated counterparts \(\{F(y)\}\), while  \(D_{\text{\scriptsize Y}}\) focuses on discriminating between \(\{y\}\) and \(\{G(x)\}\). 

What sets CycleGAN apart is the inclusion of cycle-consistency loss, which enforces that the translation from one domain to the other and back should bring the image back to its original state. This ensures that the generated images are coherent and realistic. So, the goal is to minimize adversarial losses, which aim to align the distribution of generated images with the data distribution in the target domain; and cycle consistency losses, implemented to avoid inconsistencies between the learned mappings \(G\) and \(F\).

Adversarial losses are applied to both mapping functions. The objective for mapping \(G\) and its discriminator \(D_{\text{\scriptsize Y}}\) is formulated as:

\begin{equation}
\begin{aligned}
\mathcal{L}_{\text{GAN}}(G, D_Y, X, Y) = & \mathbb{E}_{y \sim p_{\text{data}}(y)}[\log D_Y(y)] \\
& + \mathbb{E}_{x \sim p_{\text{data}}(x)}[\log(1 - D_Y(G(x)))].
\end{aligned}
\end{equation}

Where generator tries to generate images that are similar to images in the target domain while discriminator attempts to distinguish between generated images and real samples. Thus, it is a minimax problem. A similar adversarial loss is introduced for the mapping \(F\) and its discriminator \(D_{\text{\scriptsize X}}\).

As previously stated, it is essential for the mapping functions to be cycle-consistent. This implies that for every \textit{x} in domain \(X\), there is \(\textit{x} \rightarrow G(\textit{x}) \rightarrow F(G(\textit{x})) \approx \textit{x}\), which is referred to as forward cycle-consistency. Similarly, backward cycle-consistency is defined for domain \(Y\). So, and the cycle-consistency loss is formed by the combination of these two aspects:

\begin{equation}
\begin{aligned}
\mathcal{L}_{\text{cyc}}(G, F) = & \mathbb{E}_{x \sim p_{\text{data}}(x)}[||F(G(x)) - x||_1] \\
& + \mathbb{E}_{y \sim p_{\text{data}}(y)}[||G(F(y)) - y||_1].
\end{aligned}
\end{equation}

There is another loss term in CycleGAN, which is referred to as identity loss, designed to enhance the model's performance by preserving the essential characteristics of input data during the image translation process. Specifically, the identity loss ensures that the generator networks maintain consistency between input and output images. This is achieved by penalizing deviations from the original data, measuring the L1 norm between the generated images and their corresponding inputs:

\begin{equation}
\begin{aligned}
\mathcal{L}_{\text{identity}}(G, F) = & \mathbb{E}_{y \sim p_{\text{data}}(y)}[||G(y) - y||_1] \\
& + \mathbb{E}_{x \sim p_{\text{data}}(x)}[||F(x) - x||_1]. 
\end{aligned}
\end{equation}

The full loss function of CycleGAN is formulated as: 
\begin{equation}
\begin{aligned}
\mathcal{L}(G, F, D_X, D_Y) &= \mathcal{L}_{\text{GAN}}(G, D_Y, X, Y) \\
&+ \mathcal{L}_{\text{GAN}}(F, D_X, Y, X) \\
&+ \lambda_{\text{cyc}} \mathcal{L}_{\text{cyc}}(G, F) \\
&+ \lambda_{\text{idt}} \mathcal{L}_{\text{identity}}(G, F).
\end{aligned}
\end{equation}
Where \(\lambda_{\text{cyc}}\) and \(\lambda_{\text{idt}}\) control the importance of \(\mathcal{L}_{\text{cyc}}(G, F)\) and \(\mathcal{L}_{\text{identity}}(G, F)\) terms respectively. So, the final goal translates to solve:
\begin{equation}
G^*, F^* = \arg \min_{G,F} \max_{D_X, D_Y} \mathcal{L}(G, F, D_X, D_Y).
\end{equation}

In our proposed method, we employ CycleGAN to translate synthetic images into the real domain. The CycleGAN model is trained using a synthetic dataset, which is generated with Unreal Engine following the methodology outlined in the previous section and a combination of NYU-Depth V2, House Rooms Image ~\cite{reni2023house}, House Rooms ~\cite{lu2023houserooms}, and Position Image ~\cite{position-jat94_dataset} datasets as the real-world dataset.  From these real datasets, we selectively used images containing objects or scenes that closely resemble the simulated data. The dataset created by CycleGAN, through its translation of images from synthetic to real domain, is subsequently used to train our depth estimation model. Figure \ref{fig:GANresults} illustrates examples of synthetic images generated using Unreal Engine alongside their translated counterparts produced by CycleGAN.

\subsection{Depth Estimation}
We implemented the DenseDepth~\cite{DenseDepth} model for depth estimation, which utilizes a sophisticated encoder-decoder architecture connected through skip connections. The encoder is based on DenseNet-169 pre-trained on ImageNet. It encodes the input RGB image into a feature vector. The decoder contains upsampling blocks consisting of 2× bilinear upsampling followed by two 3×3 convolutional layers. The convolutional layer of the decoder is skip-connected with the encoder's pooling layer which has the same spatial dimension. Our training method employs a dataset consisting of indoor scene images and their corresponding ground truth grayscale depth map. 

During the training process, the model is updated based on pixel-wise loss that integrates three components: depth loss, gradient loss, and structural similarity loss.
\begin{equation}
\begin{aligned}
L(y, \hat{y}) &= \lambda_{\text{depth}} L_{\text{depth}}(y, \hat{y}) + L_{\text{grad}}(y, \hat{y}) + L_{\text{SSIM}}(y, \hat{y}) 
\end{aligned}
\end{equation}

The first loss term \( L_{\text{depth}} \) employs a point-wise L1 loss to minimize the absolute difference between predicted and actual depth values.
\begin{equation}
L_{\text{depth}}(y, \hat{y}) = \frac{1}{n} \sum_{p}^{n} \left| y_p - \hat{y}_p \right|
\end{equation}

The second loss term \( L_{\text{grad}} \) focuses on the depth image gradients \( g \) to preserve edge information. 
\begin{equation}
L_{\text{grad}}(y, \hat{y}) = \frac{1}{n} \sum_{p}^{n} \left| g_x(y_p, \hat{y}_p) \right| + \left| g_y(y_p, \hat{y}_p) \right|
\end{equation}

Lastly, The \( L_{\text{SSIM}} \) uses structural similarity term that are commonly used metric for image reconstruction to enhance similarity between the predicted and ground truth depth maps.
\begin{equation}
L_{\text{SSIM}}(y, \hat{y}) = \frac{1 - \text{SSIM}(y, \hat{y})}{2}
\end{equation}

The sum of these loss functions enables the model to achieve high depth accuracy and the preservation of structural details.

\begin{figure}[!t]
    \centerline{\includegraphics[width=\linewidth]{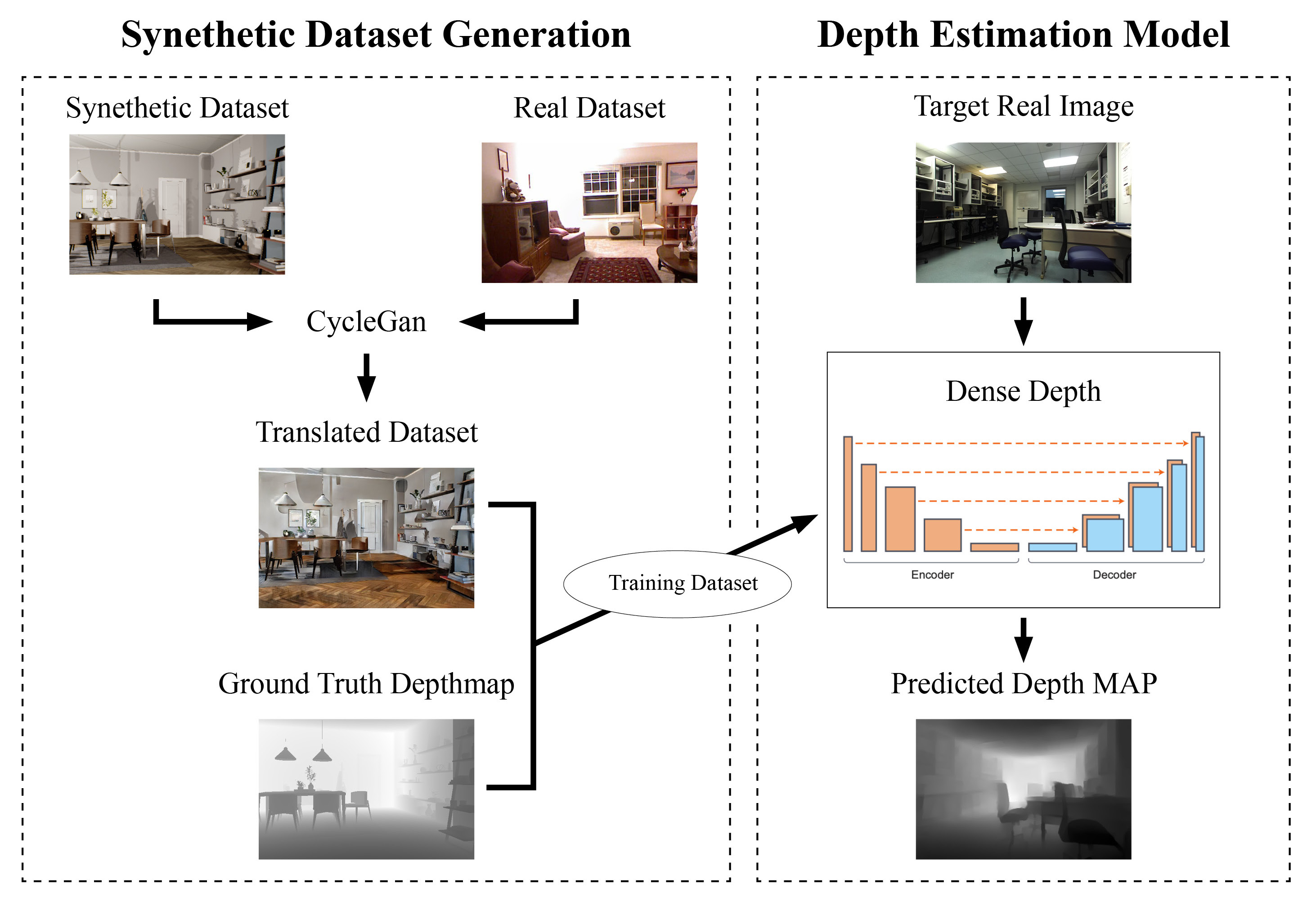}}
    \caption{We employ Unreal Engine as a synthetic data generator and DenseDepth as a depth prediction model. The generated synthetic data are translated into images similar to real-world scenes through CycleGAN. The depth prediction model is trained with the translated data.}
    \label{fig:model_architecture}
\end{figure}

\subsection{System Integration}
After the CycleGAN training and the preparation of the training data were completed, we integrated the Synthetic Data Generation and Depth Estimation modules shown in Figure \ref{fig:model_architecture} by matching up the CycleGAN-transformed images with their original ground-truth depth maps, and using the result as a training dataset for the depth estimation model. This process aligns the dimensions of the synthetic dataset that are generated by CycleGAN with the input training images for the depth estimation model. Furthermore, to achieve training dataset diversity to improve performance of depth estimation model, we randomly selected a varied set of images from the dataset. This selection aims to provide a assessment of the depth estimation model's performance across a wide range of scenarios.

%% file: 4.Experiment.tex
\section{Experimental Results}

\subsection{Dataset Composition and Pre-Processing}
To determine the impact of different types of training data on the depth prediction accuracy of the depth estimation algorithm, we created three datasets of different image compositions.
The first dataset, NYU-10k, was comprised of 10k pairs of RGB and grayscale images from the NYU-Depth V2 dataset, and was used as a comparison benchmark for the other datasets. The second dataset, UE-10k, consisted of 10k Unreal Engine simulated images from seven different environments. The third dataset, GAN-10k, consisted of the 10k simulated images from the UE-10k dataset, that had been translated to the real domain using CycleGAN. In order to further explore the impact of different types of training data, we also trained models first on the two synthetic datasets (UE-10k and GAN-10k), and then fine-tuned them by training on the NYU-10k dataset. We include the results of all five dataset combinations in our results.

\subsection{Evaluation Data Collection Using Robot}

\begin{figure}[b]
    \centerline{\includegraphics[width=\linewidth]{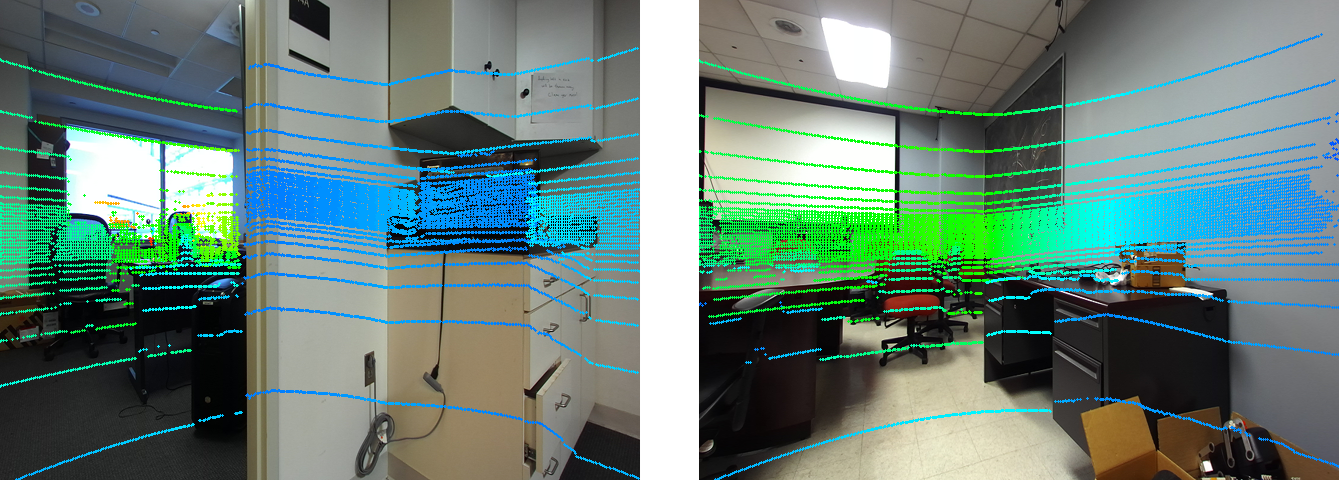}}
    \caption{Examples of LiDAR and image data collected by the Husky robot. While the density of the collected LiDAR points is not enough to fully match a predicted depth image from the network, it is enough to effectively gauge the overall accuracy of depth predictions for a given image.}
    \label{fig:lidar_example}
\end{figure}

\begin{figure}[t]
    \centerline{\includegraphics[width=\linewidth]{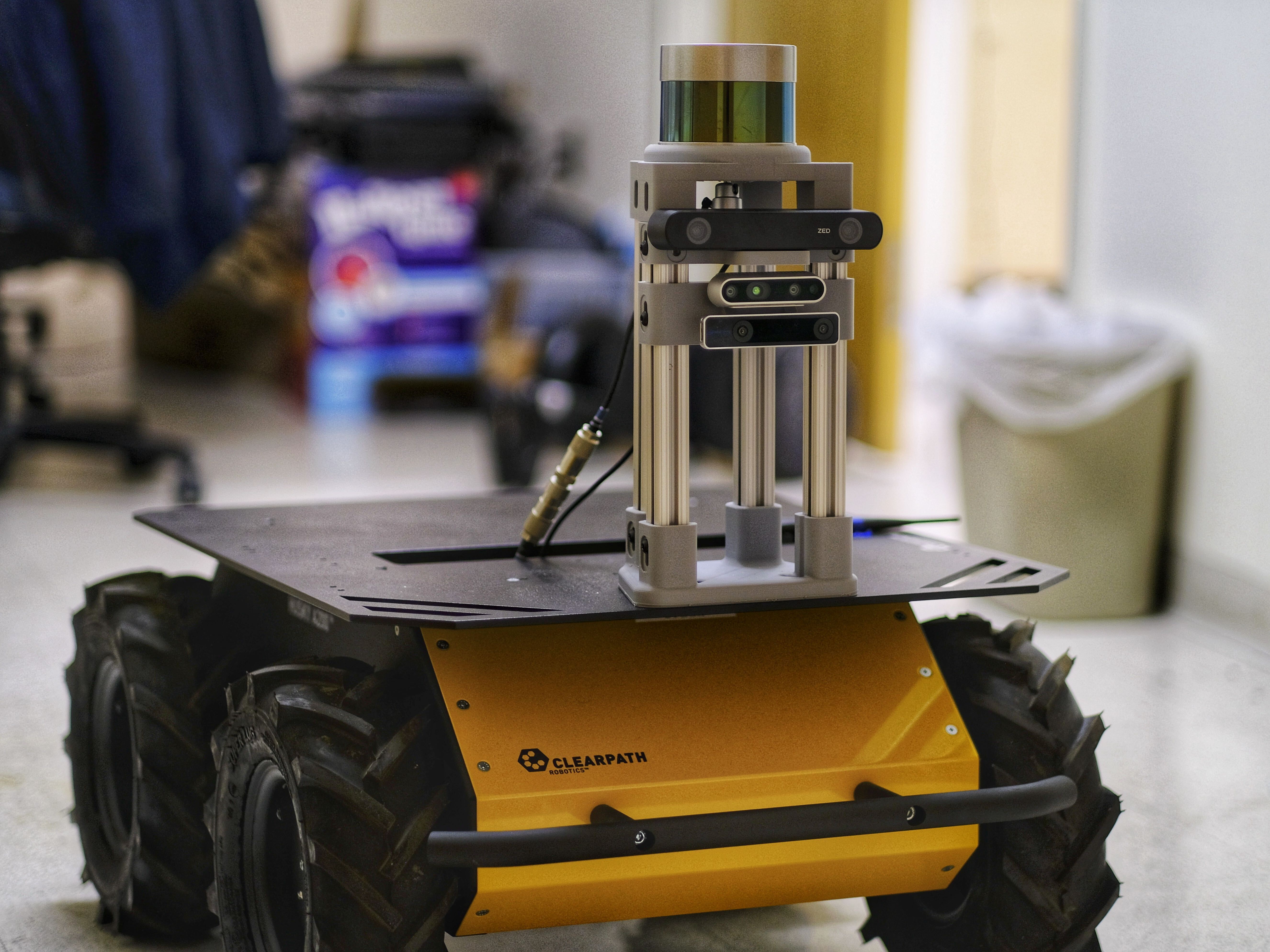}}
    \caption{ClearPath Husky robot with ZED 2 camera and Velodyne VLP-32 LiDAR unit.}
    \label{fig:husky_robot}
\end{figure}

To ensure the accuracy of our depth estimation model, we employed a Clearpath Husky robot outfitted with a ZED 2 camera and a Velodyne VLP-32 LiDAR unit, as shown in Fig.~\ref{fig:husky_robot}, to collect precise ground truth data within various environments. The robot operates on an NVIDIA Jetson Orin NX, utilizing the Robot Operating System (ROS) for navigation and data management. This platform allows the robot to autonomously navigate while capturing and recording sensor data, leveraging the ROS navigation package alongside SLAM algorithms to map and traverse the environment efficiently.

The Husky robot's data acquisition system is designed to capture comprehensive environmental data. The ZED 2 camera provides high-resolution RGB images, while the VLP-32 LiDAR captures 3D point clouds of the environment with high accuracy. Data from these sensors are recorded into a ROS bag file format. A script is employed to convert the ROS bag data into RGB images and binary point cloud data into KITTI format ~\cite{r9} enabling detailed environmental analysis and model testing.

To align the LiDAR data with the RGB images captured by the ZED 2 camera, a python script that utilizes the intrinsic and extrinsic parameters of the camera, the transformation matrix between the LiDAR and the camera was developed. This script projects the LiDAR points onto the camera's image plane. The projection is achieved through the following matrix equation:

\begin{equation}
\begin{bmatrix}
u \\
v \\
d 
\end{bmatrix}
=
\begin{bmatrix}
f_x & 0 & c_x \\
0 & f_y & c_y \\
0 & 0 & 1 
\end{bmatrix}
\begin{bmatrix}
r_{11} & r_{12} & r_{13} & t_x \\
r_{21} & r_{22} & r_{23} & t_y \\
r_{31} & r_{32} & r_{33} & t_z 
\end{bmatrix}
\begin{bmatrix}
X \\
Y \\
Z \\
1
\end{bmatrix}
\end{equation}
where \(f_x\) and \(f_y\) are the focal lengths of the camera's lens in the x and y directions, \(c_x\) and \(c_y\) are the optical center coordinates of the camera, and the rotation matrix \(r_{ij}\) and translation vector \(t_x, t_y, t_z\) define the transformation from the LiDAR frame to the camera frame.

\begin{figure*}[t]
    \centering
    \centerline{\includegraphics[width=\textwidth]{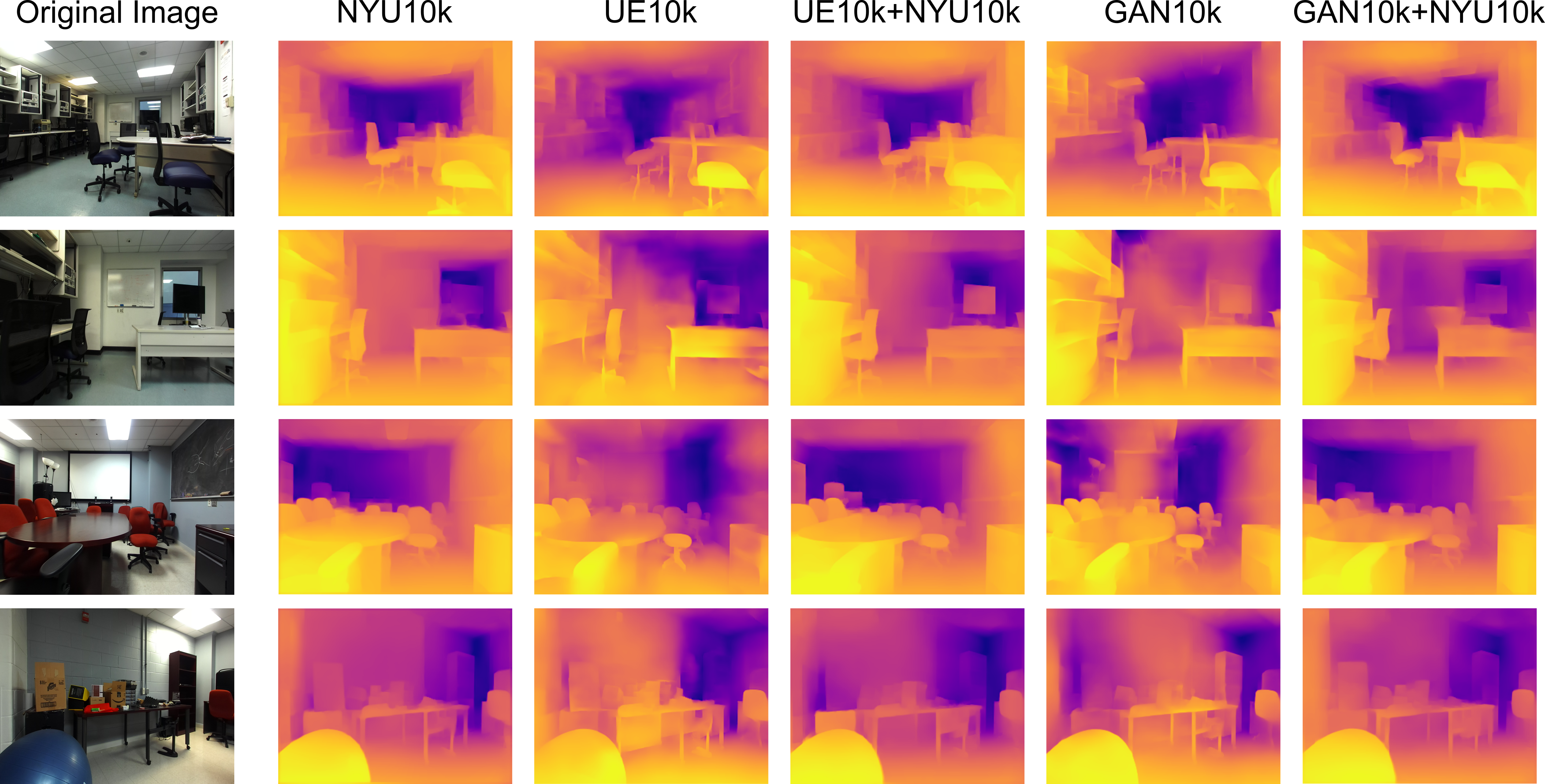}}
    \caption{Qualitative comparison between predicted depth maps: The composition of training dataset are labeled on the top. 'NYU10k' is based on the NYU Depth Dataset V2; 'UE10k' on Unreal Engine simulations; 'UE10k+NYU10k' on a mix of both; 'GAN10k' on CycleGAN-transformed images; 'GAN10k+NYU10k' on a blend of CycleGAN images and NYU dataset.}
    \label{fig:qualitative}
\end{figure*}

\subsection{Implementation }
The MDE model was trained on AMD Ryzen Threadripper 3960X 24-Core Processor and NVIDIA GeForce RTX 3090, using a batch size of 16 and a learning rate of 0.0001 with the ADAM optimizer. The model was trained using early-stopping with the patience set to 20 epochs. Because the early-stopping callback automatically stopped training after the validation accuracy stopped improving, each dataset was trained for a different number of total epochs. The model trained on NYU-10k trained for 154 epochs, the model trained on UE-10k trained for 270 epochs, and the model trained on GAN-10k trained for 181 epochs. For the combined models, the model originally trained on the UE-10k dataset was then fine tuned on the NYU-10k dataset for an additional 148 epochs, and the model originally trained on the GAN-10k dataset was fine tuned on the NYU-10k dataset for an addition 89 epochs.

For the domain transfer component of our proposed method, we implemented the CycleGAN framework as per the original specifications outlined in ~\cite{CycleGAN2017, isola2017image}. The CycleGAN was trained with the same CPU and GPU that are used training MDE model. It was trained for 200 epochs with learning rate of 0.0001.

\begin{table*}[ht]
\renewcommand{\arraystretch}{1.3}
\centering
\caption{Relative error for models using different training datasets when applied to Husky LiDAR data (first column). In the cases of the UE-10k+NYU-10k and the GAN-10k+NYU-10k datasets, the "+" indicates that model was initially trained on the first dataset and then fine tuned on the second dataset. }
\label{table:rel_error}
\begin{tabular}{|c||c|c|c|c|c|}
\hline
\textbf{LiDAR Dataset} & \textbf{NYU-10k} & \textbf{UE-10k} & \textbf{UE-10k + NYU-10k} & \textbf{GAN-10k} & \textbf{GAN-10k + NYU-10k}\\
\hline
Conference Room & 71.18\% & 73.47\% & 81.55 & \textbf{70.39\%} & 70.42\% \\
\hline
Lab Room & \textbf{51.81\%} & 59.63\% & 59.81\% & 60.72\% & 57.43\% \\	
\hline
Office & 87.48\% & 87.12\% & 107.44\% & \textbf{86.65\%} & 95.59\% \\
\hline
Lounge-Seating & 110.78\% & 96.96\% & 143.22\% & \textbf{83.92\%} & 122.01\% \\
\hline
Lounge-Kitchen & 89.69\% & 59.80\% & 86.63\% & \textbf{52.28\%} & 99.82\% \\
\hline
\end{tabular}
\end{table*}

\subsection{Evaluation of Depth Prediction}
To assess the accuracy and generalizability of depth models trained on datasets of varying composition, we conducted tests across different environments using newly collected LiDAR and image data gathered by a Husky robot. This evaluation aims to understand the impact of incorporating simulation data into the training set on the model's applicability in real-world scenarios.

Before conducting analysis with the collected LiDAR and image data, the images were first manually filtered to remove results that would negatively impact the evaluation. Because the robot vibrated significantly when turning quickly, the results that were removed were mainly blurry images that did not match LiDAR readings. Additionally, during testing we discovered that the mesh-backed office chairs that were present in some of the data were prone to bad LiDAR readings, resulting in inaccurate error metrics, so images where the office chair mesh took up a large portion of the image were also discarded. After removing these images, the final sizes of the Husky LiDAR test datasets were as follows: Conference Room - 124 images, Lab Room - 195 images, Office - 80 images, Lounge-Seating - 245 images, Lounge-Kitchen - 479 images.

In order to see how the different training datasets affected the applicability of the model, we compared the predicted depth map of newly collected Husky RGB images and corresponding LiDAR cloud points. The LiDAR distance data was preprocessed to truncate depth values exceeding 10 meters and rescaled to an 8-bit depth value ranging from 0 to 255. The LiDAR points are projected on the predicted depth images. Since the resolution of the LiDAR is much less than the predicted depth images, we first select each LIDAR point and then find the corresponding grayscale depth value from the predicted depth image. The relative distance error is then calculated with the two values.  This error was calculated for all LiDAR points, and the mean relative distance error was used as a metric to evaluate model performance.

The average relative distance error is defined as follows:
\begin{equation}
\text{(rel)}: \frac{1}{n} \sum_{p}^{n} \left| \frac{y_p - \hat{y}_p}{y_p} \right|\\
\end{equation}
where $y_p$ is pixel in depth image and $\hat{y}_p$ is a pixel in the predicted depth images. $n$ is the total number of pixels in the input image.

\subsection{Test Results}

Table \ref{table:rel_error} shows the relative error results for the model trained on each of the five datasets when tested on the different Husky LiDAR data sets. For the Conference Room LiDAR dataset, the model trained on the GAN-10k dataset had the best results, with the model that was pre-trained on the GAN-10k dataset and then fine-tuned on the NYU-10k dataset reaching a similar relative error. The two datasets using the original Unreal Engine data also performed well, with the model trained purely on the NYU-10k dataset having the worst results. For the Lab Room LiDAR dataset, the model trained on the NYU-10k dataset had the best results, with the other four models achieving markedly worse results. For the Office LiDAR dataset, the GAN-10k dataset again had the best results, with the NYU-10k and UE-10k datasets also achieving competitive values. In this case, the two combined datasets performed the worst. For both the Lounge-Seating and Lounge-Kitchen LiDAR datasets the model trained on the GAN-10k dataset again performed the best, with the model trained on the UE-10k dataset achieving the next best performance.

The value of the GAN based domain transfer is clearly demonstrated by comparing the results of training by the GAN domain transferred data of UE-10k versus the training with the UE-10k data without the domain transfer. In four out of five tests using the LiDAR test data, GAN transformed data resulted improved results.

Because the NYU dataset was originally collected using a Kinect camera, its depth information is limited to approximately 3-5 meters. The Lab environment was a more constrained space than the other environments, which would result in closer range values, and may explain why the models trained using the NYU-10k dataset performed well in that environment but not in the others.

Figure \ref{fig:qualitative} shows the qualitative results for the models trained on the five different dataset combinations. It is generally the case that estimating distances to closer objects resulted lower errors.  While all the datasets produced effective models, the model trained on the GAN-10k dataset produced more accurate details, especially in the mid-ground of images.

\subsection{Limitations}

\begin{figure}[t]
    \centerline{\includegraphics[width=\linewidth]{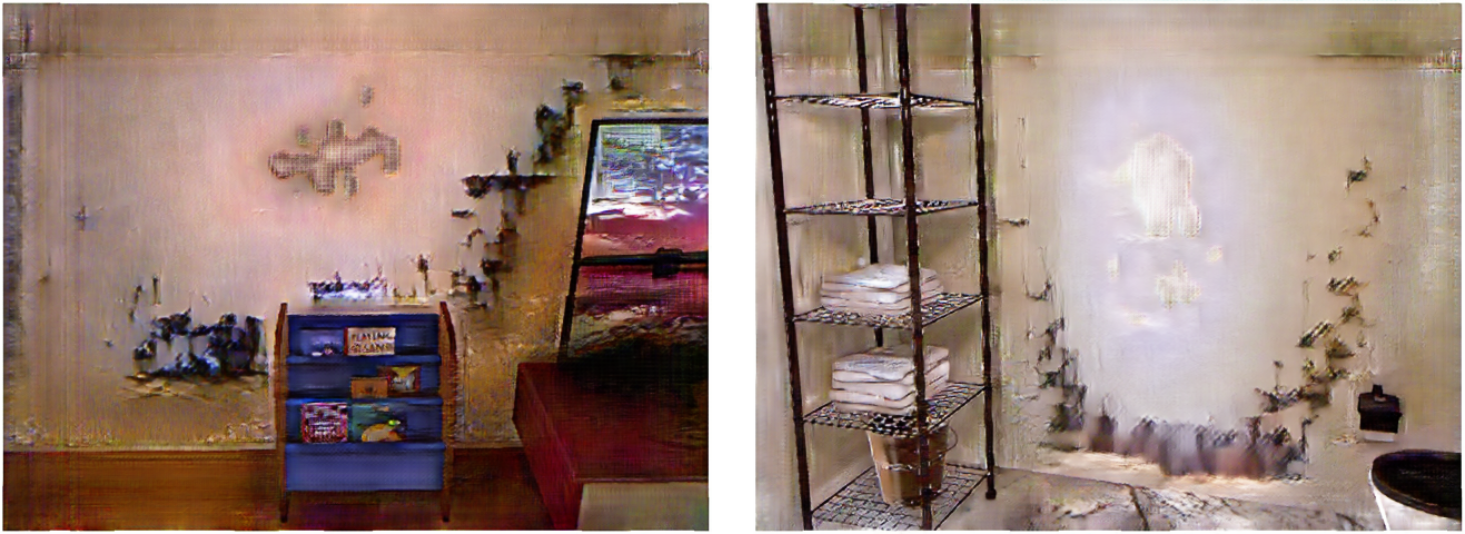}}
    \caption{Examples of artifacts present in some CycleGAN-transformed images. While the overall image structure remains intact, some parts of the image, especially blank surfaces, can contain hallucinated textures or colors.}
    \label{fig:gan_artifacts}
\end{figure}

While we showed that GAN-transformed data can serve as a suitable alternative to real-world data, there are several improvements that could be made to both the data creation process and to the validation experiments. One factor that may have negatively impacted the results of the CycleGAN model was the inclusion of translated images with artifacts in the dataset. Because CycleGAN transformation is not a perfect process, the results can sometimes contain artifacts that do not correspond to the original input image. Figure \ref{fig:gan_artifacts} shows examples of artifacts that can occur in GAN-translated images. This was the case for some of the data used in the GAN-10k dataset, which likely impacted the results. While we chose to include those images in this work in order to preserve dataset size and diversity, future work could benefit by defining criteria to filter out images with translation artifacts.

Additionally, the synthetic environments used to generate the simulated data were mainly limited to indoor home environments, which could have constrained the results. Including more environments with a wider range of object details and lighting conditions could have a positive impact on the resulting models, and make them more robust to varying conditions.

The amount of available validation data was constrained by two factors; the available collection environments and the need for an accurate transformation between the LiDAR sensor and camera. Specifically, depending on the floor surface of the room where data was collected, the Husky robot would shake drastically when turning, making the resulting LiDAR transformation unreliable and the data unusable. This could be overcome by securing more data collection locations with smoother floors, or by switching to a robot with a different drive system.

%% file: 5.Conclusion.tex
\section{Conclusion}

In this work we proposed a method for improving simulated data for use with monocular depth estimation models through the use of CycleGAN domain transfer. By generating images with depth information in a 3D simulation environment and utilizing CycleGAN domain transfer to increase their realism, we showed that synthetic datasets can achieve accuracy that is competitive with models trained using real data. Through the use of real-world experiments using a Husky robot and collected LiDAR data, we verified the performance of models trained with different dataset composition, and showed that the model trained on the GAN-10k dataset performed the best on four of the five LiDAR evaluation datasets.

Synthetic data offers many advantages over real data, such as lower collection costs, higher resolution depth information, unlimited range capability, and more, but was often limited by data realism. By using CycleGAN domain transfer to help bridge that gap, synthetic data can become a powerful tool for training effective monocular depth estimation models.